\documentclass[letterpaper]{article} 
\usepackage[]{aaai25}  
\usepackage{times}  
\usepackage{helvet}  
\usepackage{courier}  
\usepackage[hyphens]{url}  
\usepackage{graphicx} 
\usepackage{natbib}  
\usepackage{caption} 
\usepackage{newfloat}

\title{Per-Domain Generalizing Policies: On Validation Instances and Scaling Behavior
}
\author{
   Timo P. Gros\textsuperscript{\rm 1,2},
   Nicola J. Müller\textsuperscript{\rm 1,2},
   Daniel Fišer\textsuperscript{\rm 3},
   Isabel Valera\textsuperscript{\rm 1},
   Verena Wolf\textsuperscript{\rm 1,\rm 2},
   J\"{o}rg Hoffmann\textsuperscript{\rm 1,\rm 2}
}
\affiliations {
   \textsuperscript{\rm 1}Saarland University, Saarland Informatics Campus, Saarbr\"{u}cken, Germany \\
   \textsuperscript{\rm 2}German Research Center for Artificial Intelligence (DFKI), Saarbr\"{u}cken, Germany \\
   \textsuperscript{\rm 3}Aalborg University, Denmark \\
   \{timopgros,nmueller,ivalera,wolf,hoffmann\}@cs.uni-saarland.de, danfis@danfis.cz
}

\usepackage{bibentry}

\usepackage{xcolor,colortbl}
\usepackage[mathscr]{euscript}
 \let\mathscr\relax
\usepackage{amsthm}
\usepackage{amsmath}
\usepackage{amssymb}
\usepackage{calc}
\usepackage{amsfonts}
\usepackage{mathrsfs} 
\usepackage{stmaryrd}
\usepackage{mathtools}
\usepackage{siunitx}
\usepackage{harpoon}
\usepackage{bbold}
\usepackage{wasysym}
\usepackage{multirow}
\usepackage{caption}
\usepackage{subcaption}
\usepackage{diagbox}
\usepackage[shortlabels]{enumitem}
\usepackage{xspace}
\usepackage{etoolbox}
\usepackage{ifthen}
\usepackage{booktabs}
\usepackage{nicematrix}
\usepackage{amssymb}

\ifdefined\nocflagdefined

\newcommand{\nicola}[1]{}
\newcommand{\joerg}[1]{}
\newcommand{\danielh}[1]{}
\newcommand{\danf}[1]{}
\newcommand{\verena}[1]{}
\newcommand{\timo}[1]{}
\newcommand{\review}[1]{}
\else

\newcommand{\nicola}[1]{\textcolor{red}{NICOLA: \textbf{#1}}}
\newcommand{\joerg}[1]{\textcolor{red}{JOERG: \textbf{#1}}}
\newcommand{\danf}[1]{\textcolor{red}{DANF: \textbf{#1}}}
\newcommand{\verena}[1]{\textcolor{red}{VERENA: \textbf{#1}}}
\newcommand{\timo}[1]{\textcolor{red}{TIMO: \textbf{#1}}}
\newcommand{\review}[1]{\textcolor{blue}{REVIEW: \textbf{#1}}}

\fi

\newcommand{\nm}[1]{\nicola{1}}
\newcommand{\vw}[1]{\verena{1}}
\newcommand{\jh}[1]{\joerg{1}}
\newcommand{\df}[1]{\danf{1}}

\usepackage[linesnumbered,ruled,vlined,noend]{algorithm2e}
\SetAlFnt{\small}
\SetAlCapFnt{\small}
\SetAlCapNameFnt{\small}
\SetArgSty{textnormal}
\SetFuncSty{textnormal}
\SetFuncArgSty{textnormal}
\SetCommentSty{textnormal}
\SetNlSty{bfseries}{\color{black}}{}

\makeatletter
\newcommand{\MySetKwFunction}[2]{%
  \expandafter\gdef\csname @#1\endcsname##1{\FuncSty{#2\ensuremath{(}}\FuncArgSty{##1}\FuncSty{\ensuremath{)}}}%
  \expandafter\gdef\csname#1\endcsname{%
    \@ifnextchar\bgroup{\csname @#1\endcsname}{\FuncSty{#2}\xspace}}%
}%
\makeatother
\SetKw{inlineif}{if}
\SetKw{inlineelse}{else}
\SetKw{GoTo}{go to}
\SetKw{Continue}{continue}
\SetKw{Break}{break}
\SetKwInput{Input}{Input}
\SetKwInput{Parameters}{Parameters}
\SetKwInput{Output}{Output}
\MySetKwFunction{algoUniformRandom}{uniformRandom}
\MySetKwFunction{algoPartialRun}{partialRun}
\MySetKwFunction{algoSelect}{weightedSelect}
\SetKwFunction{algoCheckValidSize}{instanceOfSizeExists}
\SetKwFunction{algoGenerateInstances}{generateInstance}
\SetKwFunction{algoRunPolicy}{runPolicy}

\newcommand{\policy}{\ensuremath{\pi}}
\newcommand{\generator}{\ensuremath{\mathcal{G}}}
\newcommand{\coverageThreshold}{\ensuremath{\tau}}
\newcommand{\terminationThreshold}{\ensuremath{\zeta}}
\newcommand{\instanceSize}{\ensuremath{n}}
\newcommand{\insufficientCoverages}{\ensuremath{\text{fails}}}
\newcommand{\coverage}{\ensuremath{\mathcal{C}}}
\newcommand{\instances}{\ensuremath{\mathcal{I}}}
\newcommand{\results}{\ensuremath{\mathcal{R}}}
\newcommand{\validationScore}{\ensuremath{v}}

\theoremstyle{plain}

\theoremstyle{definition}

\newtheorem*{remark*}{Remark}

\newenvironment{proofsketch}{%
  \proof}{\endproof}

\definecolor{eyecancerpink}{rgb}{1.0, 0.0, 1.0}

\definecolor{aqua}{rgb}{0.0, 1.0, 1.0}
\definecolor{atomictangerine}{rgb}{1.0, 0.6, 0.4}
\definecolor{awesome}{rgb}{1.0, 0.13, 0.32}

\newcommand*{\eg}{e.g.\@\xspace}
\newcommand*{\ie}{i.e.\@\xspace}

\definecolor{eclipseBlue}{RGB}{42,0.0,255}
\definecolor{eclipseGreen}{RGB}{63,127,95}
\definecolor{eclipsePurple}{RGB}{127,0,85}
\colorlet{punct}{red!60!black}
\definecolor{delim}{RGB}{20,105,176}

\newcommand{\curly}[1]{\left\{#1\right\}}

\newboolean{isarxiv}
\setboolean{isarxiv}{true}
\ifthenelse{\boolean{isarxiv}}{\nocopyright}{}

\begin{document}

\maketitle

\begin{abstract}
Recent work has shown that successful per-domain generalizing action
policies can be learned. Scaling behavior, from small training
instances to large test instances, is the key objective; and the use
of validation instances larger than training instances is one key to
achieve it. Prior work has used fixed validation sets. Here, we
introduce a method generating the validation set dynamically, on the
fly, increasing instance size so long as informative and feasible. We
also introduce refined methodology for evaluating scaling behavior,
generating test instances systematically to guarantee a given
confidence in coverage performance for each instance size. In
experiments, dynamic validation  improves
scaling behavior of GNN policies in all $9$ domains used.
\end{abstract}

\section{Introduction}
\label{intro}



\emph{Per-domain generalization} in PDDL planning is a useful and
popular setting for learning policies. Prior work has shown that
successful policies of this kind can be learned using neural
architectures \cite[e.g.,][]{groshev:etal:icaps-18,garg:etal:icaps-19,toyer2018action,toyer:etal:jair-20,rivlin2020generalized,staahlberg2022learning,staahlberg2022blearning,staahlberg2024learning,sharma2023symnet,rossetti2024learning,wang:thiebaux:icaps-24}.
%
%
\emph{Scaling behavior}, the ability to generalize from small training
instances to large test instances, is the key objective in this
setting. Selecting the final policy based on its performance on
\emph{validation} instances, larger than the training instances, is
one key to achieve that objective.


Intuitively, the larger the size difference between training and validation instances, the better policy selection can assess scaling behavior.
However, prior work has relied on fixed validation sets, thereby limiting the  size difference.\footnote{This also pertains to works on learning per-domain heuristic
functions \cite[e.g.,][]{chen2024learning}; some works~\cite{toyer2018action,toyer:etal:jair-20} do not use validation at all.}
As we show here, one can instead dynamically generate larger
validation instances where informative and feasible. We measure
instance size in terms of the number of objects, and we fix a
size-scaling scheme and random instance generator per domain. Given
training instances of maximal size $\instanceSize_0$, we generate
validation instances starting at $\instanceSize_0 + 1$, and we keep
generating larger instances -- a fixed number for each size -- so long
as policy coverage remains informative (above a threshold). To ensure
feasibility of this process, we impose a plan length bound.

\begin{figure}[h]
\includegraphics[width=0.98\columnwidth]{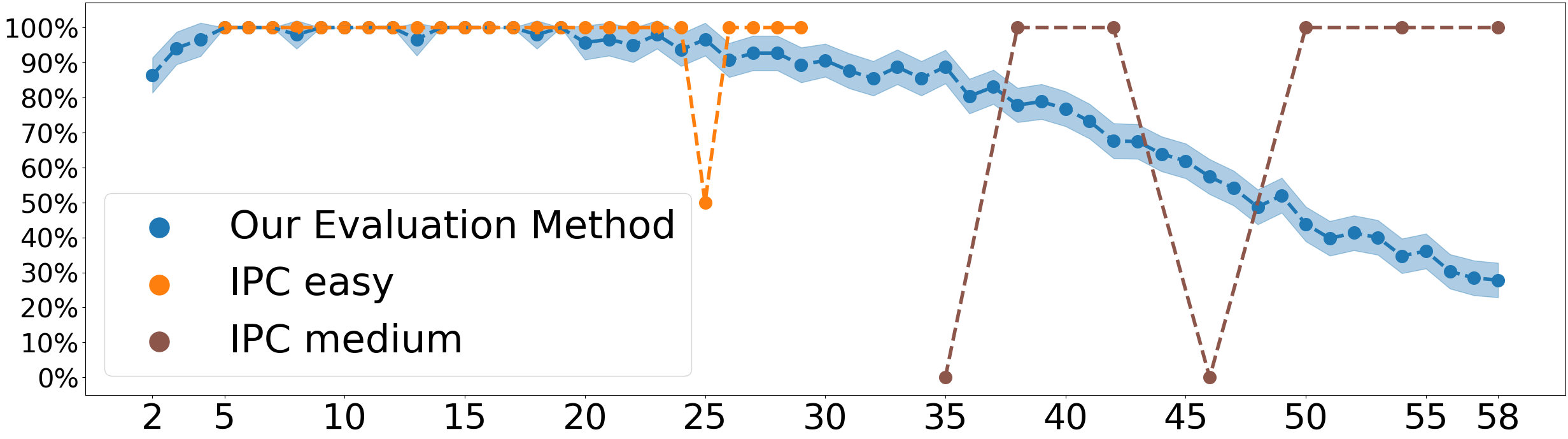}
\mbox{}
\caption{Scaling behavior (average coverage over instance size) of a
  Blocksworld policy trained following~\cite{staahlberg2022learning},
  measured on IPC'23 \cite{ayal:etal:aim-23} test sets compared to our evaluation method.}
\label{fig:intro}
\end{figure}

As an additional contribution, we introduce refined methodology for
rigorously evaluating scaling behavior in scientific
experiments. Similarly as for validation sets, prior work has used
fixed test sets, in particular ones used in International Planning
Competitions (IPCs). This is, however,
hardly adequate to meaningfully measure scaling behavior---consider
Figure~\ref{fig:intro}.

Given the non-systematic nature of IPC instance sets, there are large
gaps in instance size, and the average coverage per size mostly
trivializes to $0\%$ or $100\%$. Yet one can do much better than this,
based on the same size-scaling schemes we use for validation. For each
size value, we generate a sufficient number of instances to guarantee
a given confidence interval. The resulting plot very clearly shows how
policy performance degrades over instance size, a key insight not
visible in the IPC data at all.

We also generate the training set using the same size-scaling schemes,
giving much better justification to the i.i.d.\ assumption between
training, validation, and test data.


We run experiments with graph neural network (GNN) policies trained
following~\citet{staahlberg2022learning} on $9$ IPC'23 domains.
%
%
%
Dynamic validation consistently improves scaling behavior across all
of these domains, with substantial advantages in $8$ of them.

Our code and data are publicly available.\footnote{\url{https://doi.org/10.5281/zenodo.15314878}}

    \section{Dynamic Validation}
\label{validation}

We here introduce our dynamic validation method. We give an overview
of prior work, explain our scheme to scale instance size in a domain,
then describe the method itself.

%

\paragraph{Prior work.}

Per-domain policy learning typically relies on a form of supervised
learning~\cite{staahlberg2022learning, staahlberg2022blearning,
  staahlberg2024learning, muller2024comparing, rossetti2024learning},
training the policy to imitate an optimal planner on a set of small
training instances. To identify when the policy achieves the best
scaling behavior -- generalization to larger domain instances -- it is
validated after every epoch, assessing its current performance on a
set of larger validation instances. From all policies encountered
during this process, the one with the best validation set  performance is
selected as the final policy. Algorithm~\ref{algo:training} outlines
this training loop.

\begin{algorithm}[h]
	\Input{Training set $T$, validation set $V$, epochs $E$}
	\Output{Policy $\policy_{\text{best}}$}
	$\policy_{0} \gets$ random ; $\policy_{\text{best}} \gets \policy_{0}$ ;
	$v_{\text{best}} \gets 0$ \;
	\For{$i = 1, \dots, E$}{
	$\policy_{i} = \text{train}(\policy_{i-1}, T)$ \;
	$v_{i} = \text{validate}(\policy_{i}, V)$ \;
	\If{$v_{i} \text{ better than } v_{\text{best}}$}{
		$\policy_{\text{best}} \gets \policy_{i}$ ;
		$v_{\text{best}} \gets v_{i}$ \;
	}
	}
	\caption{\label{algo:training}Per-domain policy training loop.}
\end{algorithm}

For the validation in line $4$ of this loop, a common approach is to
compute a loss between the policy's predictions and a teacher
planner's decisions~\cite{staahlberg2022learning,
  staahlberg2022blearning, staahlberg2024learning}.  
This, however, limits the instances available for validation to only those that can be solved by a planner.
Alternatively, the policy can be validated by
running it on the validation instances and computing coverage, \ie,
the fraction of solved instances, which has the benefit of not
requiring to run the teacher planner on the validation
set~\cite{rossetti2024learning}.

All prior approaches to validation in per-domain policy learning,
to the best of our knowledge, rely on a pre-defined fixed validation
set. Yet this limits their ability to assess scaling behavior. The
data on a fixed validation set is not informative if the policy
already has perfect coverage/loss there. Further, the fixed validation
sets are typically taken from IPC instance suits, limiting the number
of available validation instances and hence the ability to see
fine-grained differences between policies. 

These limitations are quite unnecessary. As we discuss next, one can
generate validation instances on the fly, ensuring informativity for
policy selection as well as feasibility of the validation process.

\paragraph{Systematic instance size scaling.} 

%
%
Much work has been done in the past on benchmark instance scaling for
the purpose of evaluating planning systems
\cite[e.g.,][]{hoffmann:etal:jair-06,torralba:etal:icaps-21}. Here, we
merely require a systematic scheme to generate instances of scaling
size. In designing such a scheme, we stick to community conventions
and existing instance generators as much as possible.

We define instance size as the number of objects. This leaves open the
question of \emph{which} objects, i.e., given a desired size
$\instanceSize$, how to compose the object universe from the different
sub-types. Our answer is a uniform distribution over the possible
compositions given the respective IPC instance generator. Obtaining
the possible compositions is non-trivial as IPC instance generators
often do not allow to directly set the number of objects of any given
type (requiring, \eg, to instead set the x- and y-dimensions of a
map), and often implement implicit assumptions across object types
(\eg, at least one truck per city). We capture these constraints in
terms of constraint satisfaction problem
(CSP) encodings, and use a CSP solver to find valid generator
inputs that yield instances of size $\instanceSize$.

Specifically, we model instance size as a constraint $\instanceSize =
c_{1} v_{1} + \dots + c_{k} v_{k} + c_0$, where $v_{i}$ are the CSP
variables encoding the generator's arguments, and the constants $c_i$
capture the numbers of objects created by the generator. We represent
any implicit assumptions made by the generator as additional
constraints. These CSP encodings tend to be very small, and CSP
solving time is negligible.
For example, in Childsnack the task is to prepare and serve different
kinds of sandwiches to children. The generator parameters are the
number of children $v_1$, trays $v_2$, and sandwiches $v_3$. The generator
always adds $c_0=3$ tables, as well as bread and content objects for
each child yielding $c_1=3$; it returns an error if there are fewer
sandwiches than children. Accordingly, our CSP encoding is
$\instanceSize = 3v_1 + v_2 + v_3 + 3 \wedge v_1 \leq v_3$.

Another example is the Rovers domain where several rovers need to navigate between waypoints to fulfill objectives, such as gathering soil data or taking images.
The generator inputs are the number of rovers $v_1$, waypoints $v_2$, cameras $v_3$, and objectives $v_4$.
 The generator requires at least $2$ waypoints, $1$ lander object , $3$ objects representing the camera modes, and $1$ storage object for each rover.
Thus, the CSP encoding is $\instanceSize = 2 \cdot v_1 + v_2 + v_3 + v_4 + 4 \wedge 2 \leq v_2$. 

For the purpose of generating an individual instance in dynamic validation, we compute a fixed number (here set to $100$) of solutions to the CSP, sample one of these uniformly,
and pass it as input to the generator.\footnote{In scaling behavior
evaluation (see Section~\ref{evaluation}), to be even more faithful to
the generator, we compute all solutions to the CSP and sample from
these uniformly.}  If some of the generator parameters do not affect
instance size, for instance the ratio of allergic children in
Childsnack, we sample their values uniformly from the possible range.

%

\paragraph{Dynamic validation.}

Algorithm~\ref{algo:validation} outlines our dynamic validation
method.
\begin{algorithm}[h]
  \Input{Policy $\policy$, instance generator $\generator$, $CSP$, size $n_0$}
  \Parameters{per-size \#instances $m$, plan length bound $L$, coverage threshold $\coverageThreshold$}
  \Output{Validation score $\validationScore_{\pi}$}
  $\instanceSize \gets n_0 $ \;
  \Repeat{$\coverage_{\instanceSize} < \coverageThreshold$}{
	$ n \gets n + 1 $ \;
    \If{\textbf{not}\text{ }$\algoCheckValidSize(\instanceSize$)}{
    	$\coverage_n \gets 0$ ;
		\textbf{continue} \;
    }
    $\text{possibleInp} = \mathit{CSP}(\instanceSize,100) $ \;
    \For{$i \in \{1, \dots, m\}$}{
      $\text{Inp} \gets \text{uniform}(\text{possibleInp})$ \;
      $\instances \gets \algoGenerateInstances(\generator, \text{Inp})$ \;
      $\results_i \gets \algoRunPolicy(\policy, \instances, L)$ \; 
    }
    $\coverage_{\instanceSize} \gets \sum_i \results_i/m$ \;
  }
  $\validationScore_{\pi} \gets \sum_{i = n_0 + 1}^{\instanceSize} \coverage_{i}$ \;
  \caption{\label{algo:validation}Dynamic coverage validation.}
\end{algorithm}
Given training instances of maximal size $\instanceSize_0$, we
generate validation instances starting at $\instanceSize_0 + 1$. We
keep generating $m$ instances of each size so long as policy coverage
remains above a threshold $\coverageThreshold$.
To ensure feasibility of this process, we impose a plan length bound $L$.
The final validation score $\validationScore_\pi$ of policy $\pi$ is computed as the sum of achieved coverages~$\coverage_i$.

In the algorithm, we skip over values of $n$ for which no domain
instance exists according to the generator parameters and assumptions
(and hence our CSP is unsolvable). $\mathit{CSP}(\instanceSize,100)$
returns $100$ solutions to the CSP for size
$\instanceSize$. $\results_i$ is a Boolean whose value is $1$ iff the
policy found a plan.
The parameters $m$ and $\coverageThreshold$ are set manually, to a
fixed value used across all domains; in our experiments we use $m=10$
and $\coverageThreshold=30\%$. To obtain the more domain-sensitive
parameter $L$ automatically, we use $L=3N$ where $N$ is the average
length of the teacher plans on the largest training instances.


    \section{Scaling Behavior Evaluation}
\label{evaluation}

We now introduce our refined methodology for evaluating scaling
behavior encapsulated in Algorithm~\ref{algo:evaluation}.

\newcommand{\UpdateCoverage}{
	\textbf{if} $\hat\coverage_{\instanceSize} < \coverageThreshold$
	\textbf{then} $\insufficientCoverages \gets \insufficientCoverages + 1$
	\textbf{else} $\insufficientCoverages \gets 0$
}

\begin{algorithm}[h]
  \Input{Policy $\policy$, instance generator $\generator$, $\mathit{CSP}$}
  \Parameters{Statistical parameters $\epsilon$ and $\kappa$, plan length bound $L$, coverage threshold $\coverageThreshold$, consecutive fails threshold $\terminationThreshold$}
  \Output{Statistical coverage $\hat\coverage_{\instanceSize}$\ per instance size $\instanceSize$}
  $\instanceSize \gets 0$ ;
    $\insufficientCoverages \gets 0$  \;
  \While{$\insufficientCoverages < \terminationThreshold$}{
    $\instanceSize \gets \instanceSize + 1$ ; $L \gets L + 1$\;
    \lIf{\textbf{not}\text{ }$\algoCheckValidSize(\instanceSize$)}{
      \textbf{continue} 
    }
    $\hat\coverage_{\instanceSize} \gets -\infty$  ; 
    $i \gets 0$ \;
    $\text{possibleInp} = \mathit{CSP}(\instanceSize,\infty) $ \;
    \While{$P(|\hat\coverage_{\instanceSize}-\coverage_{\instanceSize}|>\epsilon)<\kappa$}{
      $\text{Inp} \gets \text{uniform}(\text{possibleInp})$ \;
      $\instances \gets \algoGenerateInstances(\generator, \text{Inp})$ \;
      $\results_i \gets \algoRunPolicy(\policy, \instances, L)$ \;
      $i \gets i + 1$ \;
      $\hat\coverage_{\instanceSize} \gets \sum_{j = 1}^i \results_j/i$ \;
    }
\UpdateCoverage
  }
  \caption{\label{algo:evaluation}Scaling behavior evaluation.}
\end{algorithm}

    

The overall mechanics of the procedure are similar to dynamic validation.
The differences are as follows. We start from $\instanceSize = 1$, so as to evaluate policy performance across the entire domain.
We again impose a plan length bound $L$ on policy executions. 
Here, we set $L = 3N + \instanceSize$, \ie, we add the current instance size, as the optimal plan length typically increases with the number of objects.
We only stop after $\terminationThreshold$ consecutive failures to meet the coverage threshold $\coverageThreshold$, to allow for temporary lapses in policy performance.
For instance generation, we draw uniformly from all possible size-$\instanceSize$ instances
($\mathit{CSP}(\instanceSize,\infty)$ returns all solutions to the CSP for size $\instanceSize$).
Instead of looking at a fixed number of instances for each size, we generate a sufficient number of instances to guarantee a given confidence interval -- precisely, with parameters $\kappa$ and $\epsilon$, we follow the Chow-Robbin's method~\cite{CR65} using a Student's t-inverval (also called sequential Student's t-inverval method): 
we keep generating runs until, with confidence $(1-\kappa)$, the half-width of the current Student's t-interval is at most $\epsilon$.
Thus, we reach a confidence of $(1-\kappa)$ that the error between average coverage $\hat\coverage_{\instanceSize}$ and real coverage $\coverage_{\instanceSize}$ is at most $\epsilon$.
We refer to the resulting value $\hat\coverage_{\instanceSize}$ as \emph{statistical
coverage}.
The algorithm outputs that value as a function of $n$.

    \section{Experiments}
\label{experiments}

\begin{figure*}[h!]
\centering
\scalebox{0.91}{
\begin{tabular}{ccc}
	\begin{subfigure}{0.33\textwidth}
		\centering
		\includegraphics[width=\textwidth]{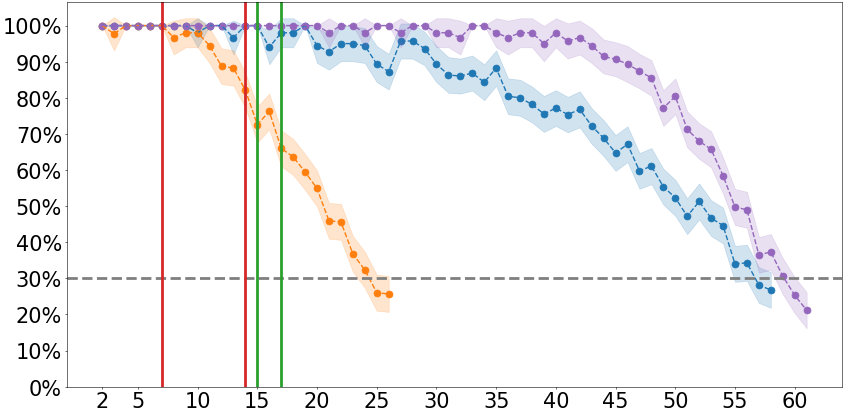}
		\caption{Blocksworld.}
	\end{subfigure} &
	\begin{subfigure}{0.33\textwidth}
		\centering
		\includegraphics[width=\textwidth]{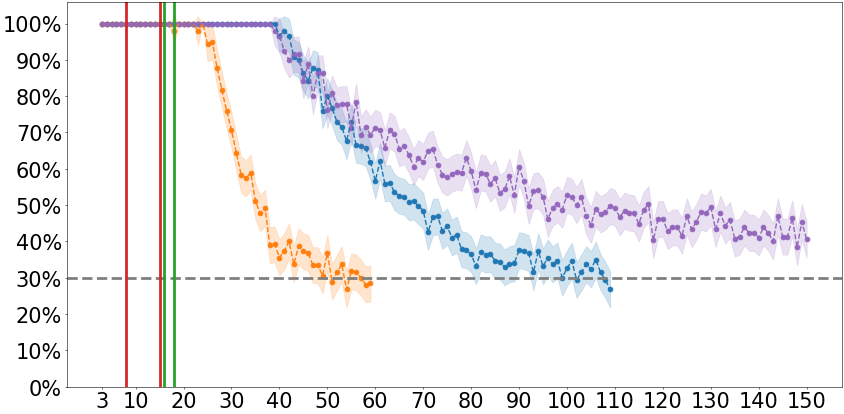}
		\caption{Ferry.}
	\end{subfigure} &
	\begin{subfigure}{0.33\textwidth}
		\centering
		\includegraphics[width=\textwidth]{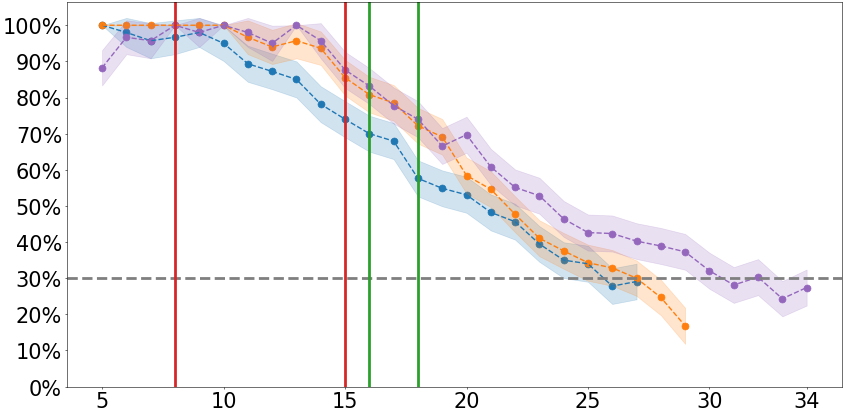}
		\caption{Satellite.}
	\end{subfigure} \\
	
	\begin{subfigure}{0.33\textwidth}
		\centering
		\includegraphics[width=\textwidth]{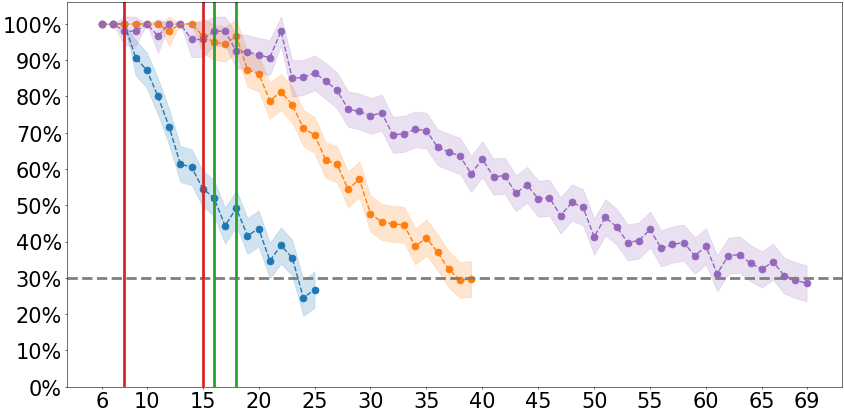}
		\caption{Transport.}
	\end{subfigure} &
	\begin{subfigure}{0.33\textwidth}
		\centering
		\includegraphics[width=\textwidth]{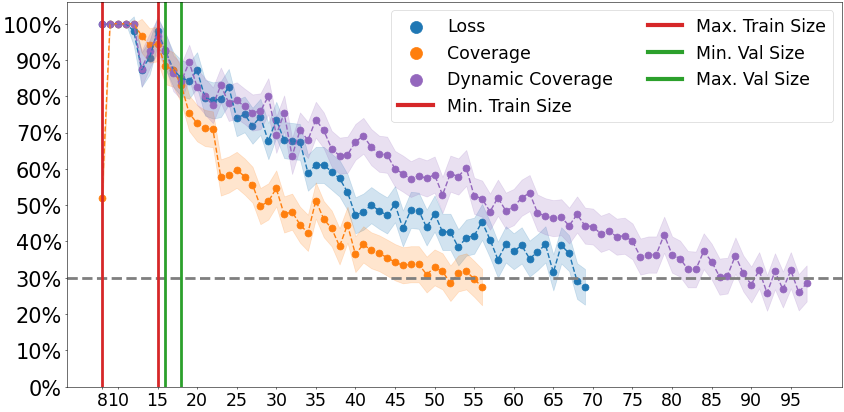}
		\caption{Childsnack.}
	\end{subfigure} &
	\begin{subfigure}{0.33\textwidth}
		\centering
		\includegraphics[width=\textwidth]{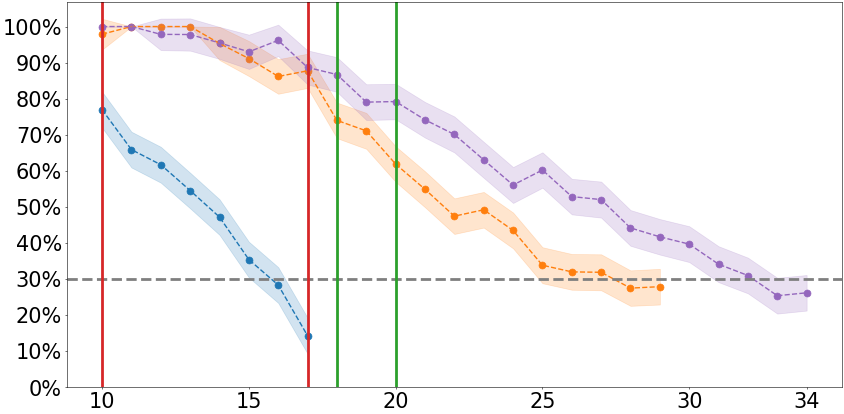}
		\caption{Rovers.}
	\end{subfigure} \\
	
	\begin{subfigure}{0.33\textwidth}
		\centering
		\includegraphics[width=\textwidth]{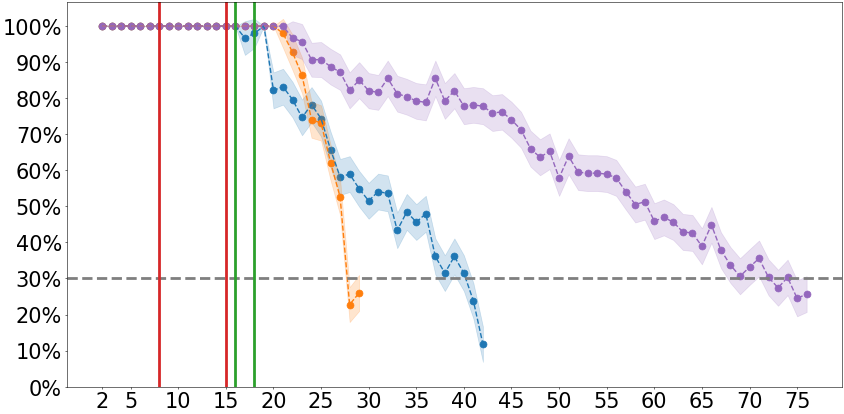}
		\caption{Gripper.}
	\end{subfigure} &
	\begin{subfigure}{0.33\textwidth}
		\centering
		\includegraphics[width=\textwidth]{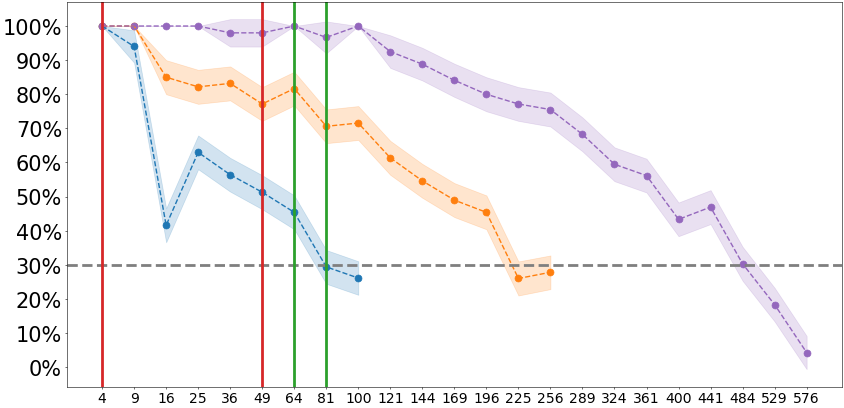}
		\caption{Visitall.}
	\end{subfigure} &
	\begin{subfigure}{0.33\textwidth}
		\centering
		\includegraphics[width=\textwidth]{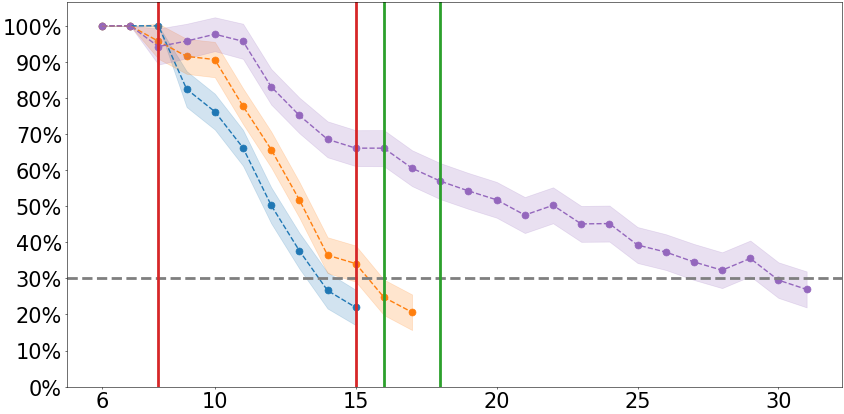}
		\caption{Logistics.}
	\end{subfigure}
\end{tabular}
}
\caption{Statistical coverage $\hat\coverage_{\instanceSize}$ over
  $\instanceSize$ of policies selected using fixed-set loss (blue),
  fixed-set coverage (orange), and dynamic coverage (purple)
  validation on $9$ domains. The instance sizes used for training are
  within the vertical red lines, the instance sizes used in fixed
  validation sets are within the two green lines. Dynamic covarage
  validation starts at the lower green line. 
  We observed that the largest validation instance sizes seen during dynamic validation are often close to those seen in evaluation. {Note that we terminated the evaluation of Ferry's dynamic coverage policy early at  size $150$, as $\hat{\coverage}$ stabilized above the threshold.}
  }
\label{fig:results}
\end{figure*}

Our experiments evaluate our dynamic validation method for GNN policies against loss-based and coverage-based validation on fixed validation sets as used in prior work.
We employ our scaling behavior
evaluation methods to obtain fine-grained comparisons.
In what
follows, we outline our benchmarks, training and validation set
construction, policy training and selection setup, and empirical
results.

%

\paragraph{Benchmarks.}

We use $9$ different domains 
which is a common number for papers on
per-domain policy learning (\eg,~\citeauthor{rivlin2020generalized} $5$~\shortcite{{rivlin2020generalized}},~\citeauthor{staahlberg2022learning} $8$~\shortcite{staahlberg2022learning}, $9$~\shortcite{staahlberg2022learning}, $10$~\shortcite{staahlberg2023learning}).
$7$ of our domains have already
been used in the context of per-domain generalization, while we
additionally use Ferry and Childsnack from the learning track of the latest IPC in 2023 \cite{ayal:etal:aim-23}. 
We did not add the IPC'23 domains Floortile (none of the trained GNNs could solve any instance), Spanner and Miconic (all learned policies always achieve 100\% coverage, making them uninteresting for our experiments), and Sokoban (its generator throws an error in case an unsolvable instance is generated, requiring to run the generator over and over again, making it unfeasible).
%
The generators were taken from IPC'$23$ where available, and otherwise
from the FF domain
collection.\footnote{\url{https://fai.cs.uni-saarland.de/hoffmann/ff-domains.html}} 


\paragraph{Instance sets.}

We use the architecture of \citet{staahlberg2022learning}. The GNN learns a
state value function and the policy is obtained by greedily following
the best action, \ie, the action leading to the state with the lowest
state value. To prevent cycles, we prohibit the policy from visiting
states more than once~\cite{staahlberg2022blearning}.
\ifthenelse{\boolean{isarxiv}}
{The training hyperparameters can be found in Appendix~\ref{hyperparameters}.}
{Details about the training hyperparameters are provided in the technical appendix~\cite{techapp}.}

We construct the training sets by uniformly sampling $100$ instances
per size, discarding duplicates. For the fixed validation sets, we
start at size $n_0 + 1$ where $n_0$ is the largest training size. Per
size, we generate $100$ instances discarding duplicates, and uniformly
select $4$ of these.
%
%
%
Full details about the training and validation sets are provided in
\ifthenelse{\boolean{isarxiv}}
{Appendix~\ref{datasets}.}
{the technical appendix~\cite{techapp}.}
We use the \textit{seq-opt-merge-and-shrink}
configuration of Fast Downward~\cite{helmert:jair-06} with limits of
$20$ minutes and $64$ GB as the teacher planner, computing optimal plans
for both sets.


\paragraph{Joint policy training and selection.}

Our setup is designed such that the training procedure is performed
jointly for all 3 validation methods. After every training epoch,
we apply each method in turn, fixed-set loss (henceforth: loss),
fixed-set coverage (henceforth: coverage), and our dynamic-set
coverage method as introduced in Section~\ref{validation} (henceforth:
dynamic coverage). At the end of training, for each validation method,
we select the respective best policy. In this manner, we guarantee
that any differences in policy performance \emph{are exclusively due
to the difference in validation methods}.

In coverage, we imposed the same plan length limit as in dynamic
coverage. We also imposed a one-hour time limit on the validation
processes, but this was never reached in our experiments.
In our experiments, dynamic coverage validation took about $8$ times as long as coverage validation. 
However, this overhead only occurs during training.

\paragraph{Results.}

Figure~\ref{fig:results} shows the scaling behavior evaluation of the
policies validated based on loss (blue), coverage (orange), and
dynamic coverage (purple). 

The policies selected by the two fixed-set validation methods (loss and
coverage) have mixed performances, which are similar only in the Satellite, Childsnack, and Logistics domains.
Dynamic validation, however, consistently yields the best scaling behavior across all domains.
Table~\ref{tab:results} provides a summary view of these results.
\begin{figure*}
	\centering
	\scalebox{0.91}{
		\begin{tabular}{ccc}
			\begin{subfigure}{0.33\textwidth}
				\centering
				\includegraphics[width=\textwidth]{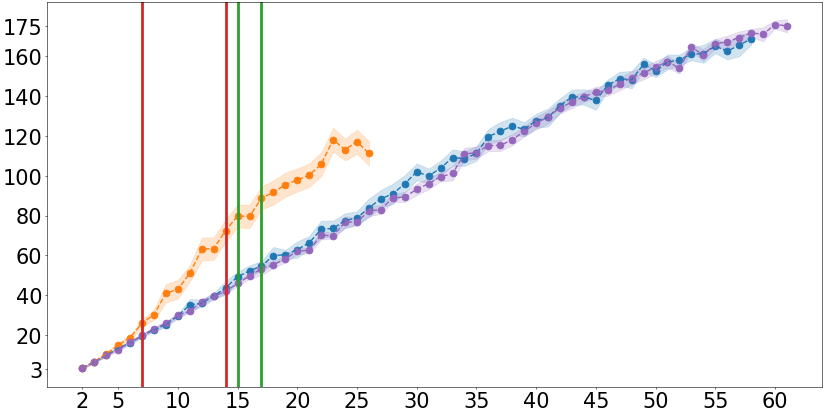}
				\caption{Blocksworld.}
			\end{subfigure} &
			\begin{subfigure}{0.33\textwidth}
				\centering
				\includegraphics[width=\textwidth]{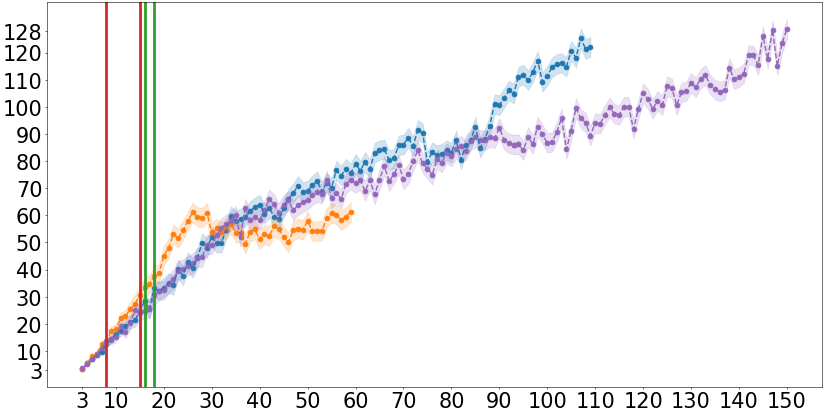}
				\caption{Ferry.}
			\end{subfigure} &
			\begin{subfigure}{0.33\textwidth}
				\centering
				\includegraphics[width=\textwidth]{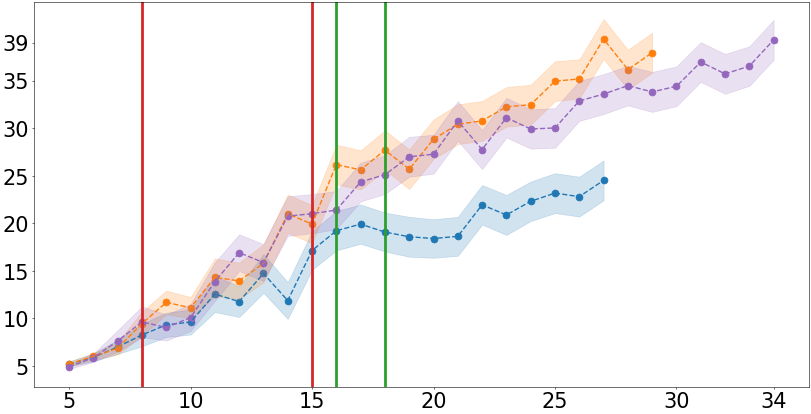}
				\caption{Satellite.}
			\end{subfigure} \\
			
			\begin{subfigure}{0.33\textwidth}
				\centering
				\includegraphics[width=\textwidth]{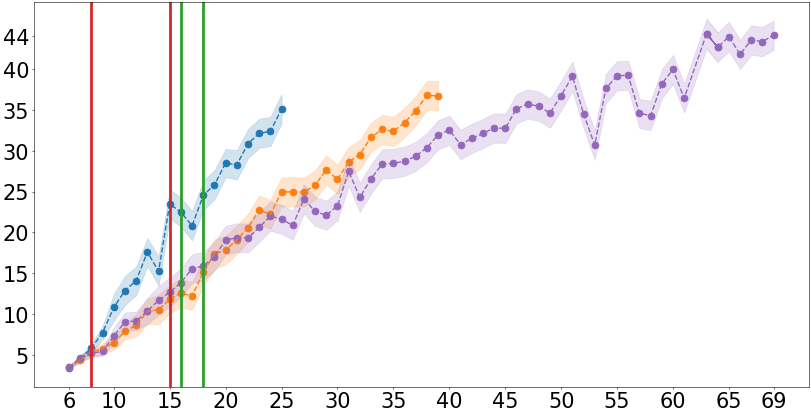}
				\caption{Transport.}
			\end{subfigure} &
			\begin{subfigure}{0.33\textwidth}
				\centering
				\includegraphics[width=\textwidth]{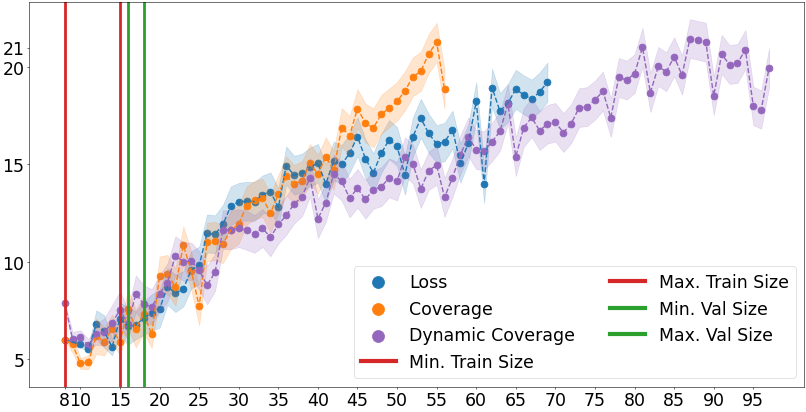}
				\caption{Childsnack.}
			\end{subfigure} &
			\begin{subfigure}{0.33\textwidth}
				\centering
				\includegraphics[width=\textwidth]{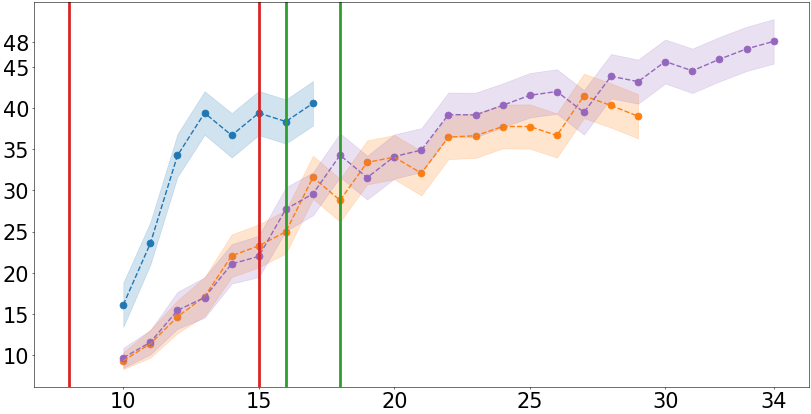}
				\caption{Rovers.}
			\end{subfigure} \\
			
			\begin{subfigure}{0.33\textwidth}
				\centering
				\includegraphics[width=\textwidth]{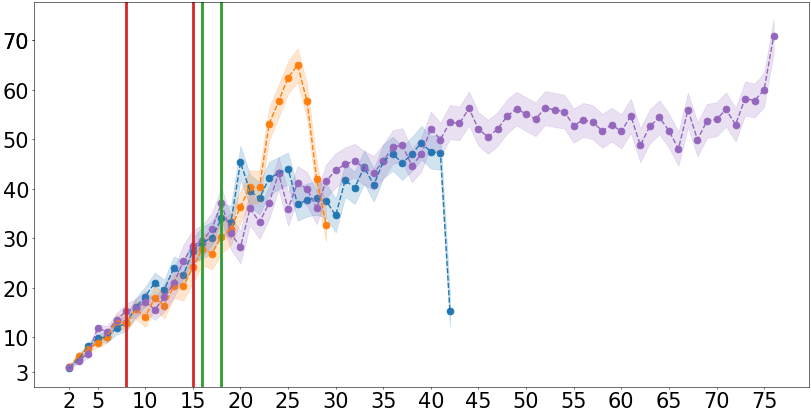}
				\caption{Gripper.}
			\end{subfigure} &
			\begin{subfigure}{0.33\textwidth}
				\centering
				\includegraphics[width=\textwidth]{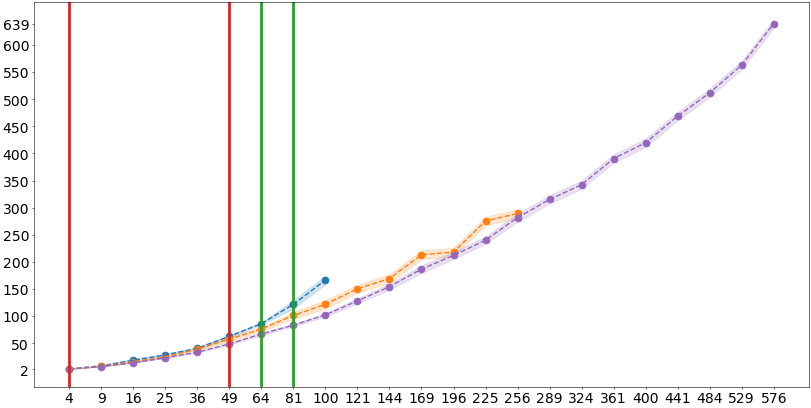}
				\caption{Visitall.}
			\end{subfigure} &
			\begin{subfigure}{0.33\textwidth}
				\centering
				\includegraphics[width=\textwidth]{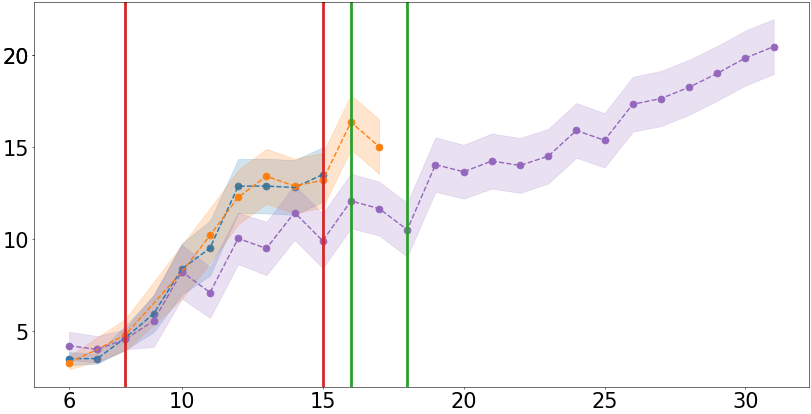}
				\caption{Logistics.}
			\end{subfigure}
		\end{tabular}
	}
	\caption{Evaluation of plan length using the same policies and instance sizes as in Figure~\ref{fig:results}.}
	\label{fig:plan_len_results}
\end{figure*} 
\begin{table}
	\renewcommand{\arraystretch}{1.2}
	\centering
	\scalebox{0.8}{
	\begin{tabular}{| l | >{\centering\arraybackslash}p{0.6cm} >{\centering\arraybackslash}p{1.1cm} | >{\centering\arraybackslash}p{0.6cm} >{\centering\arraybackslash}p{1.1cm} | >{\centering\arraybackslash}p{0.6cm} >{\centering\arraybackslash}p{1.1cm} |}
		\hline
		 & \multicolumn{2}{c|}{\textbf{Loss}} & \multicolumn{2}{c|}{\textbf{Coverage}} & \multicolumn{2}{c|}{\textbf{Dynamic}} \\
		\textbf{Domain} & Scale & SumCov   & Scale & SumCov  & Scale & SumCov \\
		\hline
		Blocksworld & 56 & 45.42 & 24 & 17.97 & \textbf{59} & \textbf{52.25} \\ 
		\hline
		Ferry & 107 & 72.24 & 57 & 37.74 & \textbf{148} & \textbf{99.34} \\ 
		\hline
		Satellite & 25 & 15.03 & 27 & 17.03 & \textbf{32} & \textbf{19.33} \\ 
		\hline
		Transport & 23 & 11.46 & 37 & 24.0 & \textbf{67} & \textbf{41.25} \\ 
		\hline
		Childsnack & 67 & 37.06 & 54 & 26.45 & \textbf{95} & \textbf{51.59} \\ 
		\hline
		Rovers & 15 & 3.41 & 27 & 12.57 & \textbf{32} & \textbf{16.32} \\ 
		\hline
		Gripper & 40 & 29.84 & 27 & 24.39 & \textbf{74} & \textbf{53.98} \\ 
		\hline
		Visitall & 64 & 4.81 & 196 & 9.88 & \textbf{484} & \textbf{17.13} \\ 
		\hline
		Logistics & 13 & 6.13 & 15 & 7.44 & \textbf{29} & \textbf{15.33} \\ 
		\hline
	\end{tabular}
} 
	\caption{Scale and SumCov scores of policies selected using
          loss, coverage, and dynamic coverage validation.
          }
	\label{tab:results}
\end{table}
We measure scaling behavior here in two ways:
\begin{itemize}
\item \emph{Scale} is the
largest instance size $n$ before the policy falls below the threshold
$\coverageThreshold$ for $\terminationThreshold$ consecutive times,
measuring how far the policy can generalize with a sufficient performance; and
\item \emph{SumCov} sums up statistical coverage up to instance size
$n$, measuring the ``area below the coverage curve''.
\end{itemize}

Our dynamic validation method performs best in all domains in both
measures:
The \emph{Scale} measure shows that the policies selected by dynamic validation generalize consistently to larger instance sizes than the policies selected by loss or coverage validation.
According to the \emph{SumCov} measure, dynamic validation policies also achieve higher total coverages across all instance sizes than either the loss or the coverage validation policies.


Recall that the policy selection was performed on the same training run, so
the superiority of dynamic validation is exclusively due to policy
selection.

\paragraph{Analyzing further properties.}
So far, scaling behavior evaluation was only used to analyze coverage, as it is the main objective of per-domain generalizing policies.
However, this method can also be applied to any other trajectory-based property.
As an example, consider Figure~\ref{fig:plan_len_results}, where we analyze the average plan length (by discarding runs that timeout) of the same policies and for the same instance sizes used in Figure~\ref{fig:results}.

We observe that the policies selected using dynamic coverage not only scale to larger instance sizes but additionally find  plans of equal or even shorter length, with the sole exception of the Satellite domain.
Given that the dynamic coverage policies solve more instances, which means also harder ones where more actions are needed, this is a very promising insight.
(Another interesting comparison would be to compare the average plan length only on instances that all of the 3 approaches solve, which we leave  for future work.)





    \section{Conclusion}
\label{conclusion}

Per-domain generalization is a natural and popular setting for policy
learning. Our work contributes new insights into the use of validation
for policy selection in this context, and into the evaluation of
scaling behavior for empirical performance analysis. The results are
highly encouraging, showing improvements in all $9$
domains used.

An intriguing aspect of these improvements is that they are obtained
through \emph{policy seclection} exclusively.
Perhaps there are ways to feed back insights from validation into training, guiding the
training process towards better scaling behavior.
We note that similar approaches have been successfully employed in the context of deep reinforcement learning~\cite{gros2023dsmc, gros2024rare}.
Another interesting direction is the application of our ideas in
training processes based on reinforcement learning instead of
supervised learning
\cite[e.g.,][]{rivlin2020generalized,staahlberg2023learning}.
Since our methods are agnostic to the policy representation, all this
can in principle be done in arbitrary frameworks
\cite[e.g.,][]{toyer:etal:jair-20,rossetti2024learning}.

From the point of view of scientific experiments, an interesting
analysis could be to vary the size of the training data as well, and
determine its impact on scaling behavior. We could, for example,
examine Scale and SumCov scores as a function of training data size.

    \newpage 
    \section*{Acknowledgments}

This work was partially supported by the German Research Foundation (DFG) under grant No. 389792660, as part of TRR 248, see https://perspicuous-computing.science, by the German Federal Ministry of Education and Research (BMBF) as part of the project MAC-MERLin (Grant Agreement No. 01IW24007), by the German Research Foundation (DFG)  - GRK 2853/1 “Neuroexplicit Models of Language, Vision, and Action” - project number 471607914, and by the European Regional Development Fund (ERDF) and the Saarland within the scope of (To)CERTAIN.

    \bibliography{abbr,bib,crossref}

    \ifthenelse{\boolean{isarxiv}}
    {
    	\newpage
        \appendix
        
\section{Hyperparameters}
\label{hyperparameters}

We use similar training hyperparameters as~\citeauthor{staahlberg2022learning} with $30$ GNN layers, a hidden size of $32$, a learning rate of $0.0002$ for the Adam optimizer~\cite{kingma2014adam}, and a gradient clip value of $0.1$~\cite{staahlberg2022learning, staahlberg2022blearning}.
However, we use a fixed number of $100$ training epochs and a batch size of $1024$ for fast training.
For each domain, the training was repeated with three random seeds.

For each instance size, dynamic coverage validation generates $m=10$ instances and stops when the coverage drops below $\coverageThreshold = 30\%$. 
For scaling behavior evaluation, confidence intervals are computed with $\epsilon = 0.05$ and $\kappa = 0.1$, and the evaluation stops when the estimated coverage drops below $\coverageThreshold = 30\%$ for $\terminationThreshold = 2$ consecutive instances.

Whereas during validation we use fixed plan length bounds $L$, during scaling behavior evaluation, they are scaled linearly with the instance size.
The plan length bounds for each domain are listed in Table~\ref{tab:hyperparameters}.

\begin{center}
\begin{table}[h]
	\centering
	\begin{tabular}{|c||c|c|}
		\hline
		\textbf{Domain} & \textbf{Validation} & \textbf{Evaluation}\\ 
		\hline\hline
		Blocksworld & $120$ &  $120+\instanceSize$\\
		\hline
		Ferry & $78$ & $78+\instanceSize$ \\
		\hline
		Satellite & $39$ & $39+\instanceSize$ \\
		\hline
		Transport & $33$ & $33+\instanceSize$ \\
		\hline
		Childsnack & $15$ & $15+\instanceSize$ \\
		\hline
		Rovers & $60$ & $60+\instanceSize$ \\
		\hline
		Gripper & $60$ & $60+\instanceSize$ \\
		\hline
		Visitall & $114$ & $114+\instanceSize$ \\
		\hline
		Logistics & $27$ & $27+\instanceSize$ \\
		\hline
	\end{tabular}
	\caption{Plan length bounds used during validation and evaluation.}
	\label{tab:hyperparameters}
\end{table}
\end{center}

\newpage
\section{Datasets}
\label{datasets}

During plan generation, we discard instances for which the planner fails to find a plan within the given time and memory limits.
Additionally, we terminate the plan generation early if the planner fails to find a plan for $10$ consecutive instances.

For each domain, except Visitall, we assign the instances of the eight smallest sizes to the training set.
Similarly, for the validation set, we randomly select $12$ instances, equally distributed (if possible) among the three largest instance sizes. 
The instance sizes used for each domain are presented in Table~\ref{tab:sizes}.

We note that strictly separating instance sizes of training and validation sets is critical for generalization.
Without this separation, the policy with the best validation performance may be the one that has only learned to generalize up to the largest instance size shared between both the training and validation sets.

\begin{table}[h]
	\centering
	\begin{tabular}{|c||c|c|}
		\hline
		\textbf{Domain} & \textbf{Training} & \textbf{Validation} \\ 
		\hline\hline
		Blocksworld & $[7-14]$ & $[15-17]$ \\
		\hline
		Ferry & $[8-15]$ & $[16-18]$ \\
		\hline
		Satellite & $[8-15]$ & $[16-18]$ \\
		\hline
		Transport & $[8-15]$ & $[16-18]$ \\
		\hline
		Childsnack & $[8-15]$ & $[16-18]$ \\
		\hline
		Rovers & $[10-17] $& $[18-20]$ \\
		\hline
		Gripper & $[8-15]$ & $[16-18]$ \\
		\hline
		Visitall & $ \curly{4,9,16,\dots,49} $ & $\curly{64, 81}$ \\
		\hline
		Logistics & $[8-15]$ & $[16-18]$ \\
		\hline
	\end{tabular}
	\caption{Number of objects in instances used for training and validation sets.}
	\label{tab:sizes}
\end{table}
    }
    {}

    \end{document}